%% file: corefmain.tex
\documentclass[11pt,letterpaper]{article}
\usepackage{naaclhlt2016}
\usepackage{basecommon}
\usepackage{times}
\usepackage{latexsym} 
\usepackage{tikz}
\usepackage{url}
\usepackage{multirow}
\usepackage{adjustbox}
\usetikzlibrary{arrows,chains,matrix,positioning,scopes,fit}
\usepackage[font={small}]{caption}
\usepackage{color, colortbl}
\definecolor{Gray}{gray}{0.9}
\naaclfinalcopy 

\newcommand{\RNN}{\mathrm{\mathbf{RNN}}}

\newcommand{\ua}{\ensuremath{\mathrm{a}}}
\newcommand{\up}{\ensuremath{\mathrm{p}}}
\newcommand{\ha}{\boldh_{\ua}}
\newcommand{\hp}{\boldh_{\up}}

\renewcommand{\btheta}{\boldsymbol{\theta}}
\newcommand{\aphi}{\boldsymbol{\phi}_{\mathrm{a}}}
\newcommand{\pwphi}{\boldsymbol{\phi}_{\mathrm{p}}}
\newcommand{\ab}{\boldb_{\mathrm{\ua}}}
\newcommand{\pb}{\boldb_{\mathrm{\up}}}
\newcommand{\aW}{\boldW_{\mathrm{\ua}}}
\newcommand{\pW}{\boldW_{\mathrm{\up}}}
\newcommand{\zro}{{\color{white}0}}

\newcommand{\nicein}{\ensuremath{\,{\in}\,}}
\newcommand{\niceq}{\ensuremath{\,{=}\,}}
\newcommand{\uc}{\ensuremath{\mathrm{c}}}
\newcommand{\hc}{\boldh_{\uc}}
\newcommand{\cb}{\boldb_{\mathrm{\uc}}}
\newcommand{\cW}{\boldW_{\mathrm{\uc}}}

\newcommand{\eqpunc}[1]{{\makebox[0pt][l]{\qquad\rm{#1}}}}

\title{Learning Global Features for Coreference Resolution}

 \author{Sam Wiseman \and Alexander M. Rush \and Stuart M. Shieber \\
         School of Engineering and Applied Sciences \\ Harvard University \\ Cambridge, MA, USA \\ {\tt \{swiseman,srush,shieber\}@seas.harvard.edu }}

\date{}

\begin{document}

\maketitle

\begin{abstract}
There is compelling evidence that coreference prediction would benefit from
modeling global information about entity-clusters.
Yet, state-of-the-art performance can be achieved with systems treating each mention prediction
  independently, which we attribute to the inherent
  difficulty of crafting informative cluster-level features. We instead propose to use recurrent neural networks
  (RNNs) to learn latent, global representations of entity clusters directly
  from their mentions. We show 
  that such representations are especially useful for the prediction of pronominal mentions, and can be incorporated into
  an end-to-end coreference system that outperforms the state of the art 
  without requiring any additional search.
\end{abstract}

\section{Introduction}
\input{sections/intro}

\section{Background and Notation}
\input{sections/background}

\section{The Role of Global Features}
\label{sec:isglobalnecessary}
\input{sections/isglobalnecessary}

\section{Learning Global Features}
\label{sec:learnglobal}
\input{sections/learningglobalfeatures}

\section{Coreference with Global Features}
\label{sec:fullmod}
\input{sections/corefwithglobal}

\section{Experiments}

\subsection{Methods}
\input{sections/methods}

\subsection{Results}
\input{sections/results}

\section{Related Work}
\input{sections/relatedwork}

\section{Conclusion}
We have presented a simple, state of the art approach to incorporating global information in an end-to-end coreference system, which obviates the need to define global features, and moreover allows for simple (greedy) inference. Future work will examine improving recall, and more sophisticated approaches to global training.

\section*{Acknowledgments}
We gratefully acknowledge the support of a Google Research Award.

\nocite{koehn2004statistical}
\nocite{cort}
\bibliography{corefmain}
\bibliographystyle{naaclhlt2016}

\end{document}

%% file: sections/intro.tex
While structured, non-local coreference models would seem to hold promise for avoiding many common coreference errors (as discussed further in Section~\ref{sec:isglobalnecessary}), the results of employing such models in practice are decidedly mixed, and state-of-the-art results can be obtained using a completely local, mention-ranking system. 

In this work, we posit that global context is indeed necessary for further
improvements in coreference resolution, but argue that informative \textit{cluster}, rather than mention, level features are very difficult to devise, limiting
their effectiveness. Accordingly, we instead propose to learn representations of mention clusters by embedding them sequentially using a recurrent neural network (shown in Section~\ref{sec:learnglobal}). Our model has no manually defined cluster features, but instead learns a global representation from the individual mentions present in each cluster. We incorporate these representations into a mention-ranking style coreference system. 

The entire model, including the
recurrent neural network and the mention-ranking sub-system, is trained
end-to-end on the coreference task. We train the model as a local classifier with fixed context (that is, as a history-based model). As such, unlike several recent approaches, which may require complicated inference during training, we are able to train our model in much the same way as a vanilla mention-ranking model.

Experiments compare the use of learned global features to several
strong baseline systems for coreference resolution. We demonstrate that the
learned global representations capture important underlying information that can help
resolve difficult pronominal mentions, which remain a persistent source of  errors for modern coreference systems~\cite{DandK:13,KandK:13,wiseman15learning,martschat15latent}. Our final system improves over 0.8 points in CoNLL score over the current state of the art, and the improvement is statistically significant on all three CoNLL metrics.

%% file: sections/background.tex
Coreference resolution is fundamentally a clustering task. Given a
sequence $(x_n)_{n=1}^N$ of (intra-document) mentions -- that is, syntactic units that can refer or be referred to --
coreference resolution involves partitioning $(x_n)$ into a sequence
of clusters $(X^{(m)})_{m = 1}^M$ such that all the mentions in any
particular cluster $X^{(m)}$ refer to the same underlying
entity. Since the mentions within a particular cluster may be ordered
linearly by their appearance in the document,\footnote{We assume
  nested mentions are ordered by their syntactic heads.} we will use
the notation $X_j^{(m)}$ to refer to the $j$'th mention in the $m$'th
cluster.

A valid clustering places each mention in exactly one cluster, and so
we may represent a clustering with a vector $\boldz \nicein \{1,\ldots,M\}^N$, where $z_n \niceq m$ iff $x_n$ is a
member of $X^{(m)}$. Coreference systems attempt to
find the best clustering $\boldz^* \in \mcZ$ under some scoring
function, with $\mcZ$ the set of valid clusterings.

One strategy to avoid the computational intractability associated with predicting an entire clustering $\boldz$ is to instead predict a single \textit{antecedent} for each mention $x_n$; because $x_n$ may not be anaphoric (and therefore have no antecedents), a ``dummy'' antecedent $\epsilon$ may also be predicted. The aforementioned strategy is adopted by ``mention-ranking'' systems~\cite{DandB:08,RandN:09,DandK:13}, which, formally, predict an antecedent $\hat{y} \nicein \mcY(x_n)$ for each mention $x_n$, where $\mcY(x_n) \niceq \{1, \ldots, n-1,\epsilon\}$. Through transitivity, these decisions induce a clustering over the document.

Mention-ranking systems make their antecedent predictions with a \textit{local} scoring function $f(x_n, y)$ defined for any mention $x_n$ and any antecedent $y \nicein \mcY(x_n)$. While such a scoring function clearly ignores much structural information, the mention-ranking approach has been attractive for at least two reasons. First, inference is relatively simple and efficient, requiring only a left-to-right pass through a document's mentions during which a mention's antecedents (as well as $\epsilon$) are scored and the highest scoring antecedent is predicted. Second, from a linguistic modeling perspective, mention-ranking models learn a scoring function that requires a mention $x_n$ to be compatible with only \textit{one} of its coreferent antecedents. This contrasts with mention-pair models (e.g.,~\newcite{BandR:08}), which score all pairs of mentions in a cluster, as well as with certain cluster-based models (see discussion in \newcite{culotta2007first}). Modeling each mention as having a single antecedent is particularly advantageous for pronominal mentions, which we might like to model as linking to a single nominal or proper antecedent, for example, but not necessarily to all other coreferent mentions. 

Accordingly, in this paper we attempt to maintain the inferential simplicity and modeling benefits of mention ranking, while allowing the model to utilize global, structural information relating to $\boldz$ in making its predictions. We therefore investigate objective functions of the form 

\vspace{-5mm}
{\small
\begin{align*}
\argmax_{y_1, \ldots, y_N} \sum_{n=1}^N f(x_n, y_n) + g(x_n, y_n, \boldz_{1:n-1})\eqpunc{,}
\end{align*}
}

\vspace{-2mm}
\noindent where $g$ is a global function that, in making predictions for $x_n$, may examine (features of) the clustering $\boldz_{1:n-1}$ induced by the antecedent predictions made through $y_{n-1}$.

%% file: sections/isglobalnecessary.tex
Here we motivate the use of global features for coreference resolution by focusing on the issues that may arise when resolving pronominal mentions in a purely local way. See \newcite{clark15entity} and \newcite{stoyanov2012easy} for more general motivation for using global models.

\subsection{Pronoun Problems}
Recent empirical work has shown that the resolution of pronominal mentions accounts for a substantial percentage of the total errors made by modern mention-ranking systems. \newcite{wiseman15learning} show that on the CoNLL 2012 English development set, almost 59\% of mention-ranking precision errors and almost 24\% of recall errors involve pronominal mentions. \newcite{martschat15latent} found a similar pattern in their comparison of mention-ranking, mention-pair, and latent-tree models.  

To see why pronouns can be so problematic, consider the following passage from the ``Broadcast Conversation'' portion of the CoNLL development set (bc/msnbc/0000/018); below, we enclose mentions in brackets and give the same subscript to co-clustered mentions. (This example is also shown in Figure~\ref{fig:hidden}.)

\vspace{-1mm}
{\small
\begin{quote}
\small
\textbf{DA:} um and [I]$_1$ think that is what's - Go ahead [Linda]$_2$.

\textbf{LW:} Well and uh thanks goes to [you]$_1$ and to [the media]$_3$ to help [us]$_4$...So [our]$_4$ hat is off to all of [you]$_5$ as well.
\end{quote} 
}

\vspace{-1mm}
\noindent This example is typical of Broadcast Conversation, and it is difficult because local systems learn to myopically link pronouns such as [you]$_5$ to other instances of the same pronoun that are close by, such as [you]$_1$. While this is often a reasonable strategy, in this case predicting [you]$_1$ to be an antecedent of [you]$_5$ would result in the prediction of an incoherent cluster, since [you]$_1$ is coreferent with the singular [I]$_1$, and [you]$_5$, as part of the phrase ``all of you,'' is evidently plural. Thus, while there is enough information in the text to correctly predict [you]$_5$, doing so crucially depends on having access to the \textit{history} of predictions made so far, and it is precisely this access to history that local models lack. 

More empirically, there are non-local statistical regularities involving pronouns we might hope models could exploit. For instance, in the CoNLL training data over 70\% of pleonastic ``it'' instances and over 74\% of pleonastic ``you'' instances follow (respectively) previous pleonastic ``it'' and ``you'' instances. Similarly, over 78\% of referential ``I'' instances and over 68\% of referential ``he'' instances corefer with previous ``I'' and ``he'' instances, respectively.

Accordingly, we might expect non-local models with access to global features to perform significantly better. However, models incorporating non-local features have a rather mixed track record. For instance, \newcite{BandK:14} found that cluster-level features improved their results, whereas \newcite{martschat15latent} found that they did not. \newcite{clark15entity} found that incorporating cluster-level features \textit{beyond} those involving the pre-computed mention-pair and mention-ranking probabilities that form the basis of their agglomerative clustering coreference system did not improve performance. Furthermore, among recent, state-of-the-art systems, mention-ranking systems (which are completely local) perform at least as well as their more structured counterparts~\cite{DandK:14,clark15entity,wiseman15learning,peng15a}. 

\subsection{Issues with Global Features}
We believe a major reason for the relative ineffectiveness of global features in coreference problems is that, as noted by \newcite{clark15entity}, cluster-level features can be hard to define.  Specifically, it is difficult to define discrete, fixed-length features on clusters, which can be of variable size (or shape). As a result, global coreference features tend to be either too coarse or too sparse. Thus, early attempts at defining cluster-level features simply applied the coarse quantifier predicates \textit{all}, \textit{none}, \textit{most} to the mention-level features defined on the mentions (or pairs of mentions) in a cluster~\cite{culotta2007first,rahman11narrowing}. For example, a cluster would have the feature `most-female=true' if more than half the mentions (or pairs of mentions) in the cluster have a `female=true' feature. 

On the other extreme, \newcite{BandK:14} define certain cluster-level features by concatenating the mention-level features of a cluster's constituent mentions in order of the mentions' appearance in the document. For example, if a cluster consists, in order, of the mentions (\textit{the president}, \textit{he}, \textit{he}), they would define a cluster-level ``type'' feature `C-P-P=true', which indicates that the cluster is composed, in order, of a common noun, a pronoun, and a pronoun. While very expressive, these concatenated features are often quite sparse, since clusters encountered during training can be of any size.

%% file: sections/learningglobalfeatures.tex
\begin{figure}[t!]
\centering
\includegraphics[width=1\columnwidth]{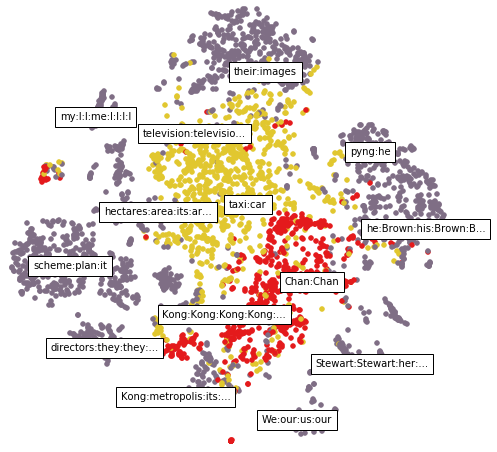}
\caption{T-SNE  visualization of
  learned entity representations on the CoNLL development set. Each point shows a gold cluster of size $>$ 1. Yellow, red, and purple points represent predominantly common
  noun, proper noun, and pronoun clusters, respectively. Captions
  show head words of representative clusters' mentions.}
\label{fig:clustviz}
\end{figure}

To circumvent the aforementioned issues with defining global features, we propose to
learn cluster-level feature representations implicitly,
by identifying the state of a (partial) cluster with the hidden state
of an RNN that has consumed the sequence of mentions composing the
(partial) cluster. Before providing technical details,
we provide some preliminary evidence that such learned
representations capture important contextual information by displaying
in Figure~\ref{fig:clustviz} the learned final states of all clusters in the
CoNLL development set, projected using
T-SNE~\cite{maaten12visualizing}. Each point in the visualization
represents the learned features for an entity cluster and the head
words of mentions are shown for representative points. Note that the
model learns to roughly separate clusters by simple distinctions such
as predominant type (nominal, proper, pronominal) and number (it, they,
etc), but also captures more subtle relationships such as grouping
geographic terms and long strings of pronouns.

\subsection{Recurrent Neural Networks}
A recurrent neural network is a parameterized non-linear function $\RNN$ that recursively maps an input sequence of vectors to a sequence of hidden states. 
Let $(\boldm_j)_{j=1}^J$ be a sequence of $J$ input vectors $\boldm_j \nicein \reals^{D}$, and let $\boldh_0 \niceq \mathbf{0}$. Applying an RNN to any such sequence yields    

\vspace{-5mm}
{\small
\begin{align*}
\boldh_j \gets \RNN(\boldm_j, \boldh_{j-1}; \btheta)\eqpunc{,}
\end{align*}
}

\vspace{-5mm}
\noindent where $\btheta$ is the set of parameters for the model, which are shared over time.

There are several varieties of RNN, but by far the most commonly used in natural-language processing is the Long Short-Term Memory network (LSTM)
\cite{hochreiter1997lstm}, particularly for language modeling (e.g., \newcite{zaremba14rnn}) and machine
translation (e.g., \newcite{sutskever2014sequence}), and we use LSTMs in all experiments. 

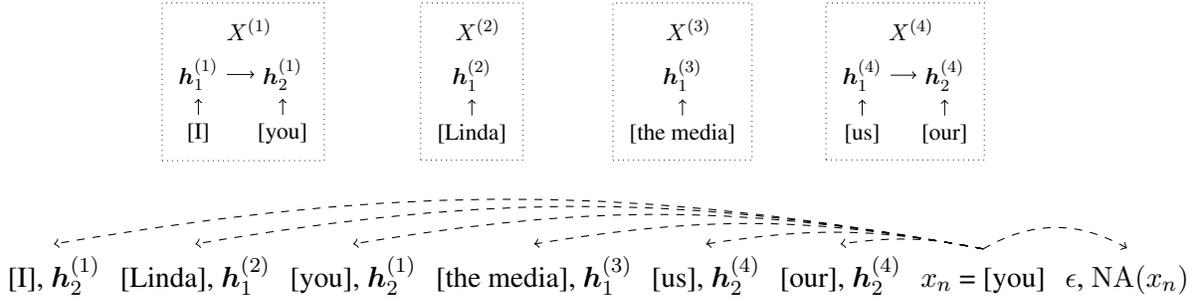
\begin{figure*}
  \centering
\vspace{-2mm}
{\small
\begin{quote}
\small
\textbf{DA:} um and [I]$_1$ think that is what's - Go ahead [Linda]$_2$.

\textbf{LW:} Well and thanks goes to [you]$_1$ and to [the media]$_3$ to help [us]$_4$...So [our]$_4$ hat is off to all of [you]$_5$...
\end{quote} 
}

\adjustbox{scale=0.8}{
\begin{tikzpicture}
    \begin{scope}[xshift=-6cm]
    \node[xshift=-0.3cm, yshift=1.2cm](X){$X^{(1)}$};
      \matrix[ nodes={ 
        line width=1pt, anchor=base, text centered, rounded corners,
        minimum width=0.5cm, minimum height=0.4mm }, row sep=0.15cm,
      column sep=0.4cm]{
        & \node(ha){$\boldh^{(1)}_1$}; & \node(hpa){$\boldh^{(1)}_2$};\\
        &  &  & & & \\
        &  \node(lexH){[I]}; & \node(toyH){[you]}; \\
      };

      \draw (ha.east) edge[->] (hpa.west); \draw (lexH) edge[->] (ha);
      \draw (toyH) edge[->] (hpa);
    \node[draw,dotted,fit=(ha) (toyH) (X)] {};
    \end{scope}

  \begin{scope}[xshift=-2cm]
    \node[xshift=-0.5cm, yshift=1.2cm](X){$X^{(2)}$};
    \matrix[ nodes={ 
      line width=1pt, anchor=base, text centered, rounded corners,
      minimum width=0.5cm, minimum height=0.4mm }, row sep=0.15cm,
    column sep=0.4cm]{
      & \node(ha){$\boldh^{(2)}_1$}; \\
      &  &  & & & \\
      &  \node(lexH){[Linda]}; \\
    };

    \draw (lexH) edge[->] (ha);
    \node[draw,dotted,fit=(ha) (lexH) (X)] {};
  \end{scope}

  \begin{scope}[xshift=1.5cm]
    \node[xshift=-0.5cm, yshift=1.2cm](X){$X^{(3)}$};
    \matrix[ nodes={ 
      line width=1pt, anchor=base, text centered, rounded corners,
      minimum width=0.5cm, minimum height=0.4mm }, row sep=0.15cm,
    column sep=0.4cm]{
      & \node(ha){$\boldh^{(3)}_1$}; \\
      &  &  & & & \\
      &  \node(lexH){[the media]}; \\
    };

    \draw (lexH) edge[->] (ha);
    \node[draw,dotted,fit=(ha) (lexH) (X)] {};
  \end{scope}

  \begin{scope}[xshift=5cm]
    \node[xshift=-0.3cm, yshift=1.2cm](X){$X^{(4)}$};
    \matrix[ nodes={ 
      line width=1pt, anchor=base, text centered, rounded corners,
      minimum width=0.5cm, minimum height=0.4mm }, row sep=0.15cm,
    column sep=0.4cm]{
      & \node(ha){$\boldh^{(4)}_1$}; & \node(hpa){$\boldh^{(4)}_2$};\\
      &  &  & & & \\
      &  \node(lexH){[us]}; & \node(toyH){[our]}; \\
    };

    \draw (ha.east) edge[->] (hpa.west); \draw (lexH) edge[->] (ha);
    \draw (toyH) edge[->] (hpa);
    \node[draw,dotted,fit=(ha) (toyH) (X)] {};
  \end{scope}
\end{tikzpicture}}
\vspace{0.25cm}

\begin{tikzpicture}[scale=0.5]
  \matrix[ nodes={ 
      line width=1pt,
      anchor=base, 
      text centered,
      rounded corners,
      minimum width=1.5cm, minimum height=8mm
    },       row sep=0.15cm]{
      \node(lex){[I], $\boldh^{(1)}_2$}; &  \node(toy){[Linda], $\boldh^{(2)}_1$};  & \node(lexb){[you], $\boldh^{(1)}_2$}; & \node(t){[the media], $\boldh^{(3)}_1$}; & \node(toyb){[us], $\boldh^{(4)}_2$}; & \node(toyc){[our], $\boldh^{(4)}_2$};  &  \node(mention){$x_n$ = [you]}; &  \node(none){$\epsilon$, $\mathrm{NA}(x_n)$};\\
};
\draw(mention.north) edge[bend right=10, ->, dashed] (lex.north);
\draw(mention.north) edge[bend right=10, ->, dashed] (toy.north);
\draw(mention.north) edge[bend right=10, ->, dashed] (lexb.north);
\draw(mention.north) edge[bend right=10, ->, dashed] (toyb.north);
\draw(mention.north) edge[bend right=10, ->, dashed] (toyc.north);
\draw(mention.north) edge[bend right=10, ->, dashed] (t.north);
\draw(mention.north) edge[bend left=30, ->, dashed] (none.north);
\end{tikzpicture} 
\caption{\small Full RNN example for handling the mention $x_n =$ [you]. There are currently four entity clusters in scope $X^{(1)}, X^{(2)}, X^{(3)}, X^{(4)}$ based on unseen previous decisions $(y)$. Each cluster has a corresponding RNN state, two of which ($\boldh^{(1)}$ and $\boldh^{(4)}$) have processed multiple mentions (with $X^{(1)}$ notably including a singular mention [I]). At the bottom, we show the complete mention-ranking process. Each previous mention is considered as an antecedent, and the global term considers the antecedent clusters' current hidden state. Selecting $\epsilon$ is treated with a special case $\mathrm{NA}(x_n)$. 
}
\label{fig:hidden} 
\end{figure*}

\subsection{RNNs for Cluster Features}
Our main contribution will be to utilize RNNs to produce
feature representations of entity clusters which will provide the basis of the global term $g$. Recall that we view a cluster $X^{(m)}$ as a sequence of mentions $(X^{(m)}_j)_{j=1}^J$ (ordered in linear document order). We therefore propose to embed the state(s) of $X^{(m)}$ by running an RNN over the cluster in order.

In order to run an RNN over the mentions we need an embedding function $\hc$ to map a mention to a real vector. First, following \newcite{wiseman15learning} define $\aphi(x_n):\mcX \rightarrow \{0,1\}^F$ as a standard set of local indicator features on a mention, such as its head word, its gender, and so on. (We elaborate on features below.) We then use a non-linear feature embedding $\hc$ to map a mention $x_n$ to a vector-space representation. In particular, we define

\vspace{-3mm}

{\small
\begin{align*}
\hc(x_n) &\triangleq \tanh(\cW \, \aphi(x_n) + \cb)\eqpunc{,}
\end{align*}
}

\vspace{-5mm}
\noindent where $\cW$ and $\cb$ are parameters of the embedding.

We will refer to the $j$'th hidden state of the RNN corresponding to $X^{(m)}$ as $\boldh^{(m)}_j$, and we obtain it according to the following formula
{\small
\begin{align*}
\boldh^{(m)}_j \gets \RNN(\hc(X_j^{(m)}), \boldh^{(m)}_{j-1}; \btheta)\eqpunc{,}
\end{align*}
} 

\vspace{-4mm}
\noindent again assuming that $\boldh^{(m)}_0 \niceq \bzero$. Thus, we will effectively run an RNN over each (sequence of mentions corresponding to a) cluster $X^{(m)}$ in the document, and thereby generate a hidden state $\boldh^{(m)}_j$ corresponding to each step of each cluster in the document.  Concretely, this can be implemented by maintaining $M$ RNNs -- one for each cluster -- that all share the parameters $\btheta$. The process is illustrated in the top portion of Figure~\ref{fig:hidden}.

%% file: sections/corefwithglobal.tex
We now describe how the RNN defined above is used within an end-to-end coreference system. 
\subsection{Full Model and Training}
\label{sec:model}
Recall that our inference objective is to maximize the score of both a local mention ranking term as well as a global term based on the current clusters: 
\vspace{-1mm}
{\small
\begin{align*}
 \argmax_{y_1,
  \ldots, y_N} \sum_{n=1}^N f(x_n, y_n) + g(x_n, y_n, \boldz_{1:n-1})
\end{align*}
}

\vspace{-3mm}
\noindent We begin by defining the local model $f(x_n,y)$ with the two layer neural network of \newcite{wiseman15learning}, which has a specialization for the non-anaphoric case, as follows:    

\vspace{-1mm}
{\small 
\begin{align*}
f(x_n,y) &\triangleq  \begin{cases} \boldu^\trans \left[ \begin{smallmatrix} \ha(x_n) \\ \hp(x_n,y) \end{smallmatrix}\right] + u_0 &\mbox{if } y \neq \epsilon \\
\boldv^\trans \ha(x_n) + v_0 &\mbox{if } y = \epsilon\eqpunc{.}  \end{cases}
\end{align*}
}

\vspace{-2mm}
\noindent Above, $\boldu$ and $\boldv$ are the parameters of the model, and $\ha$ and $\hp$ are learned feature embeddings of the local mention context and the pairwise affinity between a mention and an antecedent, respectively. These feature embeddings are defined similarly to $\hc$, as
\vspace{-1mm}
{\small
\begin{align*}
\ha(x_n) &\triangleq \tanh(\aW \, \aphi(x_n) + \ab) \\
\hp(x_n,y) &\triangleq \tanh(\pW \, \pwphi(x_n,y) + \pb)\eqpunc{,}
\end{align*}
}

\vspace{-5mm}
\noindent where $\aphi$ (mentioned above) and $\pwphi$ are ``raw'' (that is, unconjoined) features on the context of $x_n$ and on the pairwise affinity between mentions $x_n$ and antecedent $y$, respectively~\cite{wiseman15learning}. Note that $\ha$ and $\hc$ use the same raw features; only their weights differ.

We now specify our global scoring function $g$ based on the history of 
previous decisions. Define $\boldh_{<n}^{(m)}$ as the hidden state 
of cluster $m$ before a decision is made for $x_n$ -- that is, $\boldh_{<n}^{(m)}$ is the state of cluster $m$'s RNN after it has consumed all mentions in the cluster \textit{preceding} $x_n$.
We define $g$ as 
{\small\begin{align*}
g(x_n, y, &\boldz_{1:n-1}) \hspace{-4mm} &\triangleq \begin{cases} \hc(x_n)^\trans \boldh^{(z_{y})}_{<n} &\mbox{if } y \neq \epsilon \\
 \mathrm{NA}(x_n) &\mbox{if } y = \epsilon\eqpunc{,}  \end{cases}
\end{align*}}%
where $\mathrm{NA}$ gives a score for assigning $\epsilon$ based on 
a non-linear function of all of the current hidden states:

{\small\begin{align*}
  \label{eq:1}
\mathrm{NA}(x_n) =  \boldq^\trans \tanh \left( \boldW_s \left[ \begin{smallmatrix} \aphi(x_n) \\ \sum_{m=1}^{M} \boldh^{(m)}_{<n}\end{smallmatrix}\right] + \boldb_s \right).
\end{align*}}%
See Figure~\ref{fig:hidden} for a diagram. The intuition behind the first case in $g$ is that in considering whether $y$ is a good antecedent for $x_n$, we add a term to the score that examines how well $x_n$ matches with the mentions already in $X^{(z_{y})}$; this matching score is expressed via a dot-product.\footnote{We also experimented with other non-linear functions, but dot-products performed best.} In the second case, when predicting that $x_n$ is non-anaphoric, we add the NA term to the score, which examines the (sum of) the current states $\boldh^{(m)}_{<n}$ of all clusters. This information is useful both because it allows the non-anaphoric score to incorporate information about potential antecedents, and because the occurrence of certain singleton-clusters often predicts the occurrence of future singleton-clusters, as noted in Section~\ref{sec:isglobalnecessary}.

The whole system is trained end-to-end on coreference using
backpropagation. For a given training document, let $\boldz^{(o)}$
be the oracle mapping from mention to cluster, which induces an oracle
clustering.
While at training time we do have oracle clusters, we do not have oracle antecedents $(y)_{n=1}^N$, so 
following past work we treat the oracle antecedent as latent \cite{yu2009learning,fernandes2012latent,Chang:13,DandK:13}.
We train with the following slack-rescaled, margin objective:

\vspace{-5mm}
{\small\begin{align*}
\sum_{n=1}^N  \max_{\hat{y} \in \mcY(x_n)} \Delta(x_n,\hat{y}) &(1  + f(x_n,\hat{y}) + g(x_n, \hat{y}, \boldz^{(o)}) \\ 
  &- f(x_n,y_n^{\ell}) -  g(x_n, y_n^{\ell}, \boldz^{(o)})),
\end{align*}}

\vspace{-5mm}
\noindent where the latent antecedent $y_n^{\ell}$ is defined as

\vspace{-5mm}
\begin{align*}
\small
y_n^{\ell} \triangleq \argmax_{y \in \mcY(x_n): z^{(o)}_y = z^{(o)}_n} f(x_n,y) +  g(x_n, y, \boldz^{(o)}) 
\end{align*}
if $x_n$ is anaphoric, and is $\epsilon$ otherwise. The term $\Delta(x_n,\hat{y})$ gives different
weight to different error types. We
use a $\Delta$ with 3 different weights
$(\alpha_1, \alpha_2, \alpha_3)$ for ``false link'' (\textsc{fl}), ``false new'' (\textsc{fn}), and ``wrong link'' (\textsc{wl}) mistakes~\cite{DandK:13}, which correspond to predicting an antecedent when non-anaphoric, $\epsilon$ when anaphoric, and the wrong antecedent, respectively.

Note that in training we use the oracle clusters $\boldz^{(o)}$. Since these are known a priori, we can pre-compute all the hidden states $\boldh_j^{(m)}$ in a document, which makes training quite simple and efficient. This approach contrasts in particular with the work of \newcite{BandK:14} --- who also incorporate global information in mention-ranking --- in that they train against latent \textit{trees}, which are not annotated and must be searched for during training. On the other hand, training on oracle clusters leads to a mismatch between training and test, which can hurt performance.

\subsection{Search}
When moving from a strictly local objective to one with
global features, the test-time search problem becomes intractable. The
local objective requires $O(n^2)$ time, whereas the full clustering problem is NP-Hard. Past work with global features has used integer linear programming solvers for exact search
\cite{Chang:13,peng15a}, or beam search with (delayed) early update training for an
approximate solution \cite{BandK:14}. In contrast, we simply use greedy search at
test time, which also requires $O(n^2)$ time.\footnote{While beam search is a natural way to decrease search error at test time, it may fail to help if training involves a \textit{local} margin objective (as in our case), since scores need not be calibrated across local decisions. We accordingly attempted to train various locally normalized versions of our model, but found that they underperformed. 
We also experimented with training approaches and model variants that expose the model to its own predictions~\cite{daume09search,ross11a,bengio15scheduled}, but found that these yielded a negligible performance improvement.} The full algorithm is shown in Algorithm~\ref{alg:greedy}. The greedy search algorithm is identical to a simple mention-ranking system, with the exception of line~11, which updates the current RNN representation based on the previous decision that was made, and line~4, which then uses this
cluster representation as part of scoring. 
 
\begin{algorithm}[t!]
  \footnotesize
  \begin{algorithmic}[1]
    \Procedure{GreedyCluster}{$x_1, \ldots, x_N$}
    \State{Initialize clusters $X^{(1)} \ldots$ as empty lists, hidden states $\boldh^{(0)}, \ldots$ as $\mathbf{0}$ vectors in $\reals^D$, $\boldz$ as map from mention to cluster, and cluster counter {$M \gets 0$}}
    \For{$n = 2 \ldots N$ }
    \State{$\displaystyle y^* \gets \argmax_{y \in \mcY(x_n)} f(x_n, y) +  g(x_n, y, \boldz_{1:n-1}) $}
    \State{$m \gets z_{y^*}$}    
    \If{$y^* = \epsilon$}
    \State{$M \gets M + 1$}
    \State{$m \gets M$}
    \EndIf{}
    \State{append $x_n$ to  $X^{(m)}$}
    \State{$z_n \gets m$ }
    \State{$\boldh^{(m)} \gets \RNN(\hc(x_n), \boldh^{(m)})$}
    \EndFor{}
    \State{\Return{$X^{(1)}, \ldots, X^{(M)}$}}
    \EndProcedure{}
  \end{algorithmic}
  \caption{\label{alg:greedy} Greedy search with global RNNs}
\end{algorithm} 

%% file: sections/methods.tex
\begin{table*}[htpb]
\small
\centering
\begin{tabular}{lcccccccccc}
\toprule 
\multirow{2}{*}{System} & \multicolumn{3}{c}{MUC} & \multicolumn{3}{c}{B$^3$}  & \multicolumn{3}{c}{CEAF$_e$} \\
 & P & R & F$_1$ & P & R & F$_1$ & P & R & F$_1$ & CoNLL \\ 
\midrule 
B\&K (2014) & 74.3\zro & 67.46 & 70.72 & 62.71 & 54.96 & 58.58 & 59.4\zro & 52.27 & 55.61 & 61.63 \\
M\&S (2015) & 76.72 & 68.13 & 72.17 & 66.12 & 54.22 & 59.58 & 59.47 & 52.33 & 55.67 & 62.47 \\
C\&M (2015) & 76.12 & 69.38 & 72.59 & 65.64 & 56.01 & 60.44 & 59.44 & 52.98 & 56.02 & 63.02 \\
Peng et al. (2015) & - & - & 72.22 & - & - & 60.50 & - & - & 56.37 & 63.03 \\
Wiseman et al. (2015) & 76.23 & 69.31 & 72.60 & 66.07 & 55.83 & 60.52 & 59.41 & 54.88 & 57.05 &  63.39 \\
This work & 77.49 & 69.75 & \textbf{73.42} & 66.83 & 56.95 & \textbf{61.50} & 62.14 & 53.85 & \textbf{57.70} & \textbf{64.21} \\
\bottomrule 
\end{tabular} 
\caption{Results on CoNLL 2012 English test set. We compare against recent state of the art systems, including (in order) Bjorkelund and Kuhn (2014), Martschat and Strube (2015), Clark and Manning (2015), Peng et al. (2015), and Wiseman et al. (2015). F$_1$ gains are significant ($p < 0.05$ under the bootstrap resample test (Koehn, 2004)) compared with Wiseman et al. (2015) for all metrics.
}
\label{tab:mainresults}
\end{table*}

We run experiments on the CoNLL 2012 English shared task
\cite{conll12}. The task uses the OntoNotes corpus
\cite{hovy2006ontonotes}, consisting of 3,493 documents in various
domains and formats.  We
use the experimental split provided in the shared task.  For all
experiments, we use the Berkeley Coreference System \cite{DandK:13}
for mention extraction and to compute features $\aphi$ and
$\pwphi$.

\paragraph{Features} We use the raw \textsc{Basic+} feature sets described by \newcite{wiseman15learning}, with the following modifications:
\begin{itemize}
\item We remove all features from $\pwphi$ that concatenate a feature of the antecedent with a feature of the current mention, such as bi-head features.
\item We add true-cased head features, a current speaker indicator feature, and a 2-character genre (out of \{bc,bn,mz,nw,pt,tc,wb\}) indicator to $\pwphi$ and $\aphi$.
\item We add features indicating if a mention has a substring overlap with the current speaker ($\pwphi$ and $\aphi$), and if an antecedent has a substring overlap with a speaker distinct from the current mention's speaker ($\pwphi$).
\item We add a single centered, rescaled document position feature to each mention when learning $\hc$. We calculate a mention $x_n$'s rescaled document position as $\frac{2n-N-1}{N-1}$. 
\end{itemize}
These modifications result in there being approximately 14K distinct features in $\aphi$ and approximately 28K distinct features in $\pwphi$, which is far fewer features than has been typical in past work.

For training, we use document-size minibatches, which allows for efficient pre-computation of RNN states, and we minimize the loss described in Section~\ref{sec:fullmod} with AdaGrad \cite{duchi2011adaptive} (after clipping LSTM gradients to lie (elementwise) in $(-10,10)$). We find that the initial learning rate chosen for AdaGrad has a significant impact on results, and we choose learning rates for each layer out of $\{0.1,0.02,0.01,0.002,0.001\}$. 

In experiments, we set $\ha(x_n)$, $\hc(x_n)$, and $\boldh^{(m)}$ to be $\nicein \reals^{200}$, and $\hp(x_n,y) \nicein \reals^{700}$. We use a single-layer LSTM (without ``peep-hole'' connections), as implemented in the \texttt{element-rnn} library~\cite{elements}. For regularization, we apply Dropout~\cite{srivastava2014dropout} with a rate of 0.4 before applying the linear weights $\boldu$, and we also apply Dropout with a rate of 0.3 to the LSTM states before forming the dot-product scores. Following \newcite{wiseman15learning} we use the cost-weights $\boldsymbol{\alpha} = \langle 0.5, 1.2, 1 \rangle$ in defining $\Delta$, and we use their pre-training scheme as well. For final results, we train on both training and development portions of the CoNLL data. Scoring uses the official CoNLL 2012 script
\cite{pradhan2014scoring,luo2014extension}.  
Code for our system is available at \url{https://github.com/swiseman/nn_coref}. The system makes use of a GPU for training, and trains in about two hours.

%% file: sections/results.tex
\newcommand{\fix}[1]{{\color{white}1}#1}
\newcommand{\fixg}[1]{{\color{Gray}1}#1}
\begin{table}[t!]
\centering
\small
\begin{tabular}{lcccc}
\toprule 
 & MUC & B$^3$ & CEAF$_e$ & CoNLL \\ 
\midrule 
MR          & 73.06 & 62.66 & 58.98 & 64.90 \\
\rowcolor{Gray}
Avg, OH    & 73.30 & 63.06 & 58.85 & 65.07 \\
RNN, GH & 73.63 & 63.23 & 59.56 & 65.47 \\
\rowcolor{Gray}
RNN, OH  & 74.26 & 63.89 & 59.54 & 65.90 \\
\bottomrule 
\end{tabular} 
\caption{F$_1$ scores of models described in text on CoNLL 2012 development set. Rows in grey highlight models using oracle history.}
\label{tab:conlldev}
\end{table}
\begin{table}[t!]
\small
\begin{tabular}{lccc}
 \toprule 
 & \multicolumn{3}{c}{Non-Anaphoric (\textsc{fl})} \\
  & Nom.~HM & Nom.~No HM & Pron. \\
 \midrule
MR  & {1061} & \fix{130} & {1075} \\
\rowcolor{Gray}
Avg, OH  & \fixg{983} & \fixg{140} & {1011}\\
RNN, GH  & \fix{914} & \fix{125} & \fix{893}\\
\rowcolor{Gray}
RNN, OH  & \fixg{913} & \fixg{130} & \fixg{842}\\
\midrule
\# Mentions & 9.0K  & 22.2K & 3.1K \\
\toprule
 & \multicolumn{3}{c}{Anaphoric (\textsc{fn} + \textsc{wl})} \\
 Model & Nom.~HM & Nom.~No HM & Pron. \\
 \midrule
MR  & 665+326 & 666+56 & 533+796 \\
\rowcolor{Gray}
Avg, OH  & 781+300 & 641+60 & 578+744\\
RNN, GH  & 767+303 & 648+57 & 664+727 \\
\rowcolor{Gray}
RNN, OH  & 750+289 & 648+52 & 611+686 \\
\midrule
\# Mentions & 4.7K  & 1.0K & 7.3K \\
 \bottomrule
\end{tabular} 
\caption{Number of ``false link'' (\textsc{fl}) errors on non-anaphoric mentions (top) and number of ``false new'' (\textsc{fn}) and ``wrong link'' (\textsc{wl}) errors on anaphoric mentions (bottom) on CoNLL 2012 development set. Mentions are categorized as nominal or proper with (previous) head match (Nom.~HM), nominal or proper with no head match (Nom.~No HM), and pronominal. Models are described in the text, and rows in grey highlight models using oracle history.}
\label{tab:newerrors}
\end{table}

In Table~\ref{tab:mainresults} we present our main results on the CoNLL English test set, and compare with other recent state-of-the-art systems. We see a statistically significant improvement of over 0.8 CoNLL points over the previous state of the art, and the highest F$_1$ scores to date on all three CoNLL metrics. 

We now consider in more detail the impact of global features and RNNs on performance. For these experiments, we report MUC, B$^3$, and CEAF$_e$ F$_1$-scores in Table~\ref{tab:conlldev} as well as errors broken down by mention type and by whether the mention is anaphoric or not in Table~\ref{tab:newerrors}. Table~\ref{tab:newerrors} further partitions errors into \textsc{fl}, \textsc{fn}, and \textsc{wl} categories, which are defined in Section~\ref{sec:model}. We typically think of \textsc{fl} and \textsc{wl} as representing precision errors, and \textsc{fn} as representing recall errors.
 
Our experiments consider several different settings. First, we consider an oracle setting (``RNN,~OH'' in tables), in which the model receives $\boldz_{1:n-1}^{(o)}$, the oracle partial clustering of all mentions preceding $x_n$ in
the document, and is therefore not forced to rely on its own past predictions when predicting $x_n$. 
This provides us with an upper bound on the performance achievable with our model. Next, we consider the performance of the model under a greedy inference strategy (RNN,~GH), as in
Algorithm~\ref{alg:greedy}. 
Finally, for baselines we consider the mention-ranking system (MR) of \newcite{wiseman15learning} using our updated feature-set, as well as a non-local baseline with oracle history (Avg,~OH), which averages the representations $\hc(x_j)$ for all $x_j \nicein X^{(m)}$, rather than feed them through an RNN; errors are still backpropagated through the $\hc$ representations during learning.

In Table~\ref{tab:newerrors} we see that the RNN improves performance overall, with the most dramatic improvements on non-anaphoric pronouns, though errors are also decreased significantly for non-anaphoric nominal and proper mentions that follow at least one mention with the same head.
While \textsc{wl} errors also decrease for both these mention-categories under the RNN model, \textsc{fn} errors increase. Importantly, the RNN performance is significantly better than that of the Avg baseline, which barely improves over mention-ranking, even with oracle history. This suggests that modeling the sequence of mentions in a cluster is advantageous. We also note that while RNN performance degrades in both precision and recall when moving from the oracle history upper-bound to a greedy setting, we are still able to recover a significant portion of the possible performance improvement. 

\subsection{Qualitative Analysis}
In this section we consider in detail the impact of the $g$ term in the RNN scoring function on the two error categories that improve most under the RNN model (as shown in Table~\ref{tab:newerrors}), namely, pronominal \textsc{wl} errors and pronominal \textsc{fl} errors. We consider an example from the CoNLL development set in each category on which the baseline MR model makes an error but the greedy RNN model does not.

\begin{figure}[t!]
\centering
\includegraphics[width=0.85\columnwidth]{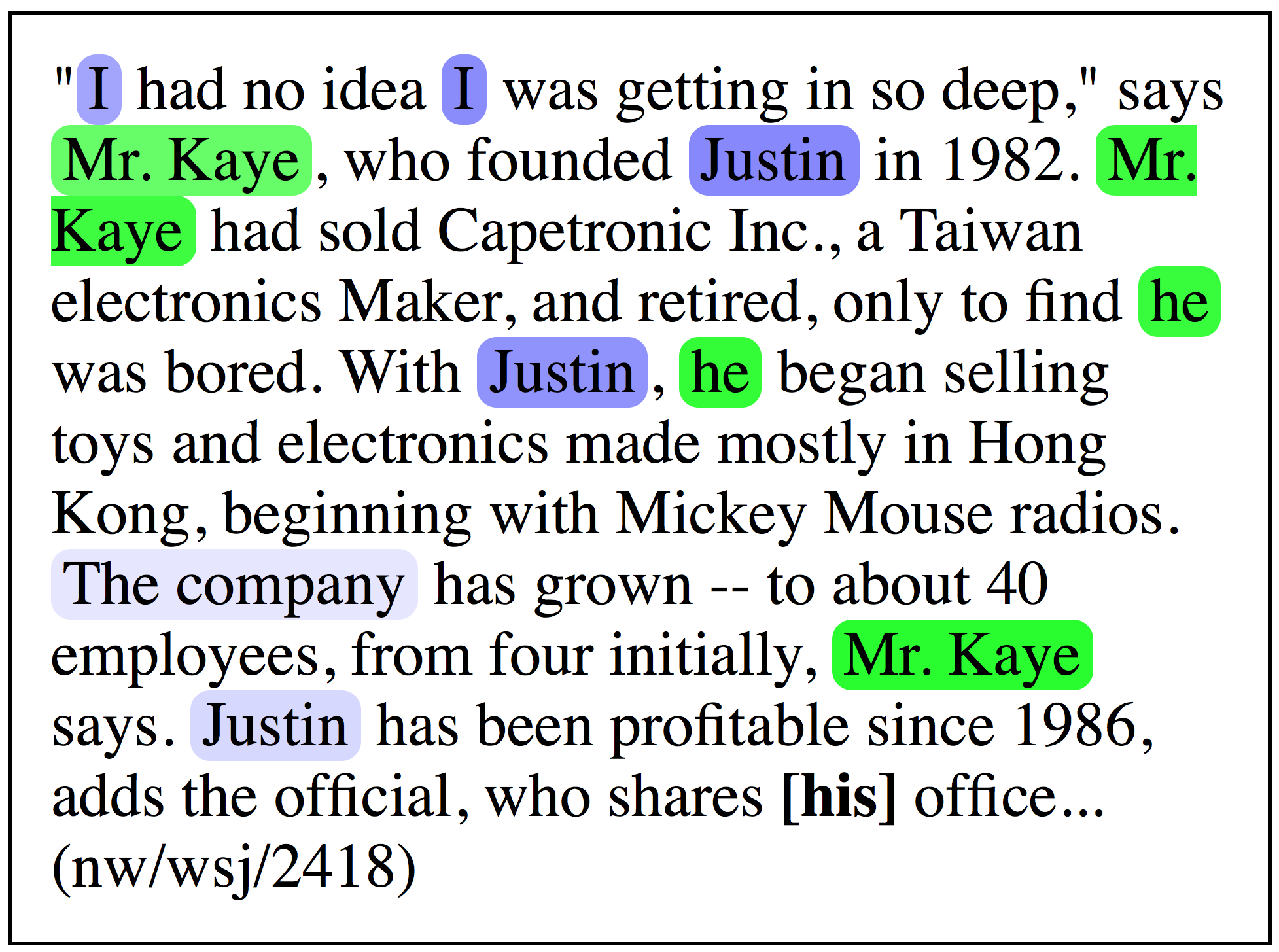}
\caption{Cluster predictions of greedy RNN model; co-clustered mentions are of the same color, and intensity of mention $x_j$ corresponds to $\boldh_c(x_n)^{\trans} \boldh_{<k}^{(i)}$, where $k \niceq j+1$, $i \in \{1,2\}$, and $x_n =$ ``his.'' See text for full description.}
\label{fig:wlviz}
\end{figure}

The example in Figure~\ref{fig:wlviz} involves the resolution of the ambiguous pronoun ``his,'' which is bracketed and in bold in the figure. Whereas the baseline MR model \textit{incorrectly} predicts ``his'' to corefer with the closest gender-consistent antecedent ``Justin'' --- thus making a \textsc{wl} error --- the greedy RNN model correctly predicts ``his'' to corefer with ``Mr. Kaye'' in the previous sentence. (Note that ``the official'' also refers to Mr. Kaye). To get a sense of the greedy RNN model's decision-making on this example, we color the mentions the greedy RNN model has predicted to corefer with ``Mr. Kaye'' in green, and the mentions it has predicted to corefer with ``Justin'' in blue. (Note that the model incorrectly predicts the initial ``I'' mentions to corefer with ``Justin.'') Letting $X^{(1)}$ refer to the blue cluster, $X^{(2)}$ refer to the green cluster, and $x_n$ refer to the ambiguous mention ``his,'' we further shade each mention $x_j$ in $X^{(1)}$ so that its intensity corresponds to $\boldh_c(x_n)^{\trans} \boldh_{<k}^{(1)}$, where $k \niceq j+1$; mentions in $X^{(2)}$ are shaded analogously. Thus, the shading shows how highly $g$ scores the compatibility between ``his'' and a cluster $X^{(i)}$ as each of $X^{(i)}$'s mentions is added. We see that when the initial ``Justin'' mentions are added to $X^{(1)}$ the $g$-score is relatively high. However, after ``The company'' is correctly predicted to corefer with ``Justin,'' the score of $X^{(1)}$ drops, since companies are generally not coreferent with pronouns like ``his.''

\begin{figure}[t!]
\centering
\includegraphics[width=0.85\columnwidth]{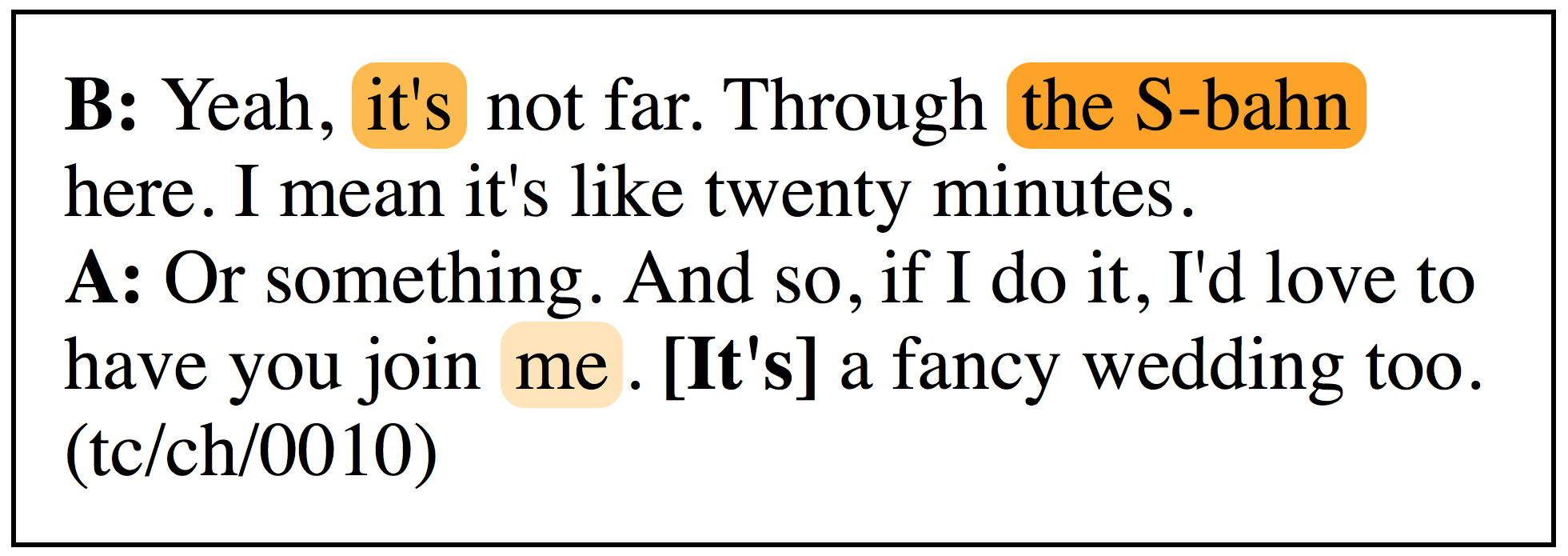}
\caption{Magnitudes of gradients of $\mathrm{NA}$ score applied to bold ``It's'' with respect to final mention in three preceding clusters. See text for full description.}
\label{fig:flviz}
\end{figure}

Figure~\ref{fig:flviz} shows an example (consisting of a telephone conversation between ``A'' and ``B'') in which the bracketed pronoun ``It's'' is being used pleonastically. Whereas the baseline MR model predicts ``It's'' to corefer with a previous ``it'' --- thus making a \textsc{fl} error --- the greedy RNN model does not. In Figure~\ref{fig:flviz} the final mention in three preceding clusters is shaded so its intensity corresponds to the magnitude of the gradient of the $\mathrm{NA}$ term in $g$ with respect to that mention. This visualization resembles the ``saliency'' technique of \newcite{li16viz}, and it attempts to gives a sense of the contribution of a (preceding) cluster in the  calculation of the $\mathrm{NA}$ score.

We see that the potential antecedent ``S-Bahn'' has a large gradient, but also that the initial, obviously pleonastic use of ``it's'' has a large gradient, which may suggest that earlier, easier predictions of pleonasm can inform subsequent predictions.

%% file: sections/relatedwork.tex
In addition to the related work noted throughout, we add supplementary references here. Unstructured approaches to coreference typically divide into mention-pair models, which classify (nearly) every pair of mentions in a document as coreferent or not~\cite{soon2001machine,ng2002identifying,BandR:08}, and mention-ranking models, which select a single antecedent for each anaphoric mention~\cite{DandB:08,RandN:09,DandK:13,Chang:13,wiseman15learning}. Structured approaches typically divide between those that induce a clustering of mentions~\cite{mccallum2003toward,culotta2007first,poon08joint,haghighi2010coreference,stoyanov2012easy,cai10end}, and, more recently, those that learn a latent tree of mentions~\cite{fernandes2012latent,BandK:14,martschat15latent}.

There have also been structured approaches that merge the mention-ranking and mention-pair ideas in some way. For instance, \newcite{rahman11narrowing} rank clusters rather than mentions; \newcite{clark15entity} use the output of both mention-ranking and mention pair systems to learn a clustering.
  
The application of RNNs to modeling (the trajectory of) the state of a cluster is apparently novel, though it bears some similarity to the recent work of \newcite{dyer15transition}, who use LSTMs to embed the state of a transition based parser's stack.